\UseRawInputEncoding
\documentclass[letterpaper, 10 pt, conference]{ieeeconf}  

\IEEEoverridecommandlockouts                              

\makeatletter
\let\NAT@parse\undefined
\makeatother

\usepackage{subfigure}
\usepackage{graphics}
\usepackage{epsfig}

\usepackage{mathptmx}
\usepackage{times}
\usepackage{amsmath}
\usepackage{amssymb}
\usepackage{float}
\usepackage{mathrsfs}
\DeclareMathAlphabet{\mathcal}{OMS}{cmsy}{m}{n}

\usepackage{amsfonts}
\usepackage{bm}
\usepackage{cite}
\usepackage{hyperref}
\usepackage{multicol}
\usepackage{multirow}
\usepackage{siunitx}
\usepackage{tikz}

\usepackage{algorithm}
\usepackage{algpseudocode}
\usepackage[linewidth=1pt]{mdframed}
\usepackage[export]{adjustbox}

\usepackage{booktabs}

\usepackage{mathtools}
\usepackage{xcolor}
\usepackage{colortbl}
\usepackage{array}
\usepackage{makecell}
\usepackage{stfloats}
\usepackage{graphicx}
\usepackage{threeparttable}

\title{\LARGE \bf
MoRI: Mixture of RL and IL Experts for Long-Horizon Manipulation Tasks
}

\author{Yaohang Xu$^{1}$, Lianjie Ma$^{1}$, Gewei Zuo$^{1}$, Wentao Zhang$^{2}$, Han Ding$^{3}$ and Lijun Zhu$^{1,\dagger}$
\thanks{This work was supported by the National Natural Science Foundation of China (U25A6013, 62173155, 52188102).}
\thanks{$^{\dagger}$Corresponding author.}
\thanks{$^{1}$School of Artificial Intelligence and Automation, Huazhong University of Science and Technology, Wuhan, China. \{yhxu, yingyi1048596, gwzuo, wentaozhang, ljzhu\}@hust.edu.cn}
\thanks{$^{2}$DRAGON Lab, The University of Tokyo, Tokyo, Japan. wentao-zhang@dragon.t.u-tokyo.ac.jp}
\thanks{$^{3}$School of Mechanical Science and Engineering, Huazhong University of Science and Technology, Wuhan, China. dinghan@hust.edu.cn}
}

\begin{document}

\maketitle
\thispagestyle{empty}
\pagestyle{empty}

\begin{abstract}
Reinforcement Learning (RL) and Imitation Learning (IL) are the standard frameworks for policy acquisition in manipulation. While IL offers efficient policy derivation, it suffers from compounding errors and distribution shift. Conversely, RL facilitates autonomous exploration but is frequently hindered by low sample efficiency and the high cost of trial and error. Since existing hybrid methods often struggle with complex tasks, we introduce Mixture of RL and IL Experts (MoRI). This system dynamically switches between IL and RL experts based on the variance of expert actions to handle coarse movements and fine-grained manipulations. MoRI employs an offline pre-training stage followed by online fine-tuning to accelerate convergence. To maintain exploration safety and minimize human intervention, the system applies IL-based regularization to the RL component. Evaluation across four complex real-world tasks shows that MoRI achieves an average success rate of 97.5\% within 2 to 5 hours of fine-tuning. Compared to baseline RL algorithms, MoRI reduces human intervention by 85.8\% and shortens convergence time by 21\%, demonstrating its capability in robotic manipulation.
\end{abstract}

\section{INTRODUCTION}
In robotic manipulation, complex tasks typically involve a combination of coarse-grained motions and contact-rich precision manipulations. A single learning paradigm often struggles to balance efficiency, accuracy, and generalization. While Imitation Learning (IL) \cite{zhao2023learningfinegrainedbimanualmanipulation} effectively acquires fundamental policies from expert demonstrations \cite{shi2023waypointbasedimitationlearningrobotic}, the resulting policy performance remains constrained by the distributional coverage of the demonstration data. This limitation results in compounding errors, which makes it struggle with the stringent requirements of high-precision manipulation \cite{10.3389/frobt.2025.1682437,simchowitz2025pitfallsimitationlearningactions}. Conversely, Reinforcement Learning (RL) \cite{10610040} optimizes policies through trial-and-error exploration and reward signals, allowing agents to perform beyond the scope of demonstration data. RL has shown significant robustness, especially in contact-rich tasks \cite{10610350,xu2026twinrlvladigitaltwindrivenreinforcement,11127855,lei2026rl100performantroboticmanipulation}. Recent progress in Human-in-the-Loop RL (HIL-RL) within real-world environments \cite{8793698,10.1126/scirobotics.ads5033,chen2025conrftreinforcedfinetuningmethod,lei2026rl100performantroboticmanipulation} has enabled the use of real-time human interventions to refine policies during on-robot deployment, leading to performance improvements in robotic tasks. Despite these gains, HIL-RL still faces inherent challenges such as slow convergence and difficulty in managing complex long-horizon tasks \cite{lu2025vlarlmasterfulgeneralrobotic,zhu2018reinforcementimitationlearningdiverse,zhou2025reinforcegenhybridskillpolicies}, which limit its effectiveness for extended or complex manipulations.

To improve sample efficiency, recent work has explored combining offline pre-training with online fine-tuning. Methods such as the Implicit Q-Learning \cite{kostrikov2021offlinereinforcementlearningimplicit} and Advantage-Weighted Actor-Critic (AWAC) \cite{nair2021awacacceleratingonlinereinforcement} use offline RL to initialize policies and then utilize advantage-weighted objectives during online interaction to accelerate policy refinement. Other studies extend consistent offline-to-online fine-tuning to Vision-Language-Action (VLA) models and real-world RL \cite{chen2025conrftreinforcedfinetuningmethod}, or employ residual RL frameworks built upon a Behavioral Cloning (BC) base policy \cite{ankile2025residualoffpolicyrlfinetuning} to facilitate learning on physical robots. Despite these advances, these approaches are often not fully validated on long-horizon tasks. While RL is frequently used for post-training VLA models \cite{chen2026pitextttrlonlinerlfinetuning,doi:10.36227/techrxiv.176531955.54563920/v1} to boost performance, the high computational cost typically hinders practical deployment. Recent efforts have also integrated iterative offline and online RL with diffusion policies \cite{lei2026rl100performantroboticmanipulation} to reduce human intervention, yet the convergence time (e.g., 14 hours per task) remains a significant bottleneck. There is still a need for an efficient framework that combines IL and RL to handle long-horizon tasks while ensuring fast convergence, minimal human intervention, and low computational overhead for both training and deployment.

To bridge the gap between IL and RL, we introduce the Mixture of RL and IL Experts (MoRI), a framework based on the Mixture-of-Experts (MoE) architecture. By leveraging the variance of expert actions, MoRI assigns coarse movements to the IL component and fine-grained manipulations to the RL component. This dynamic scheduling allows the experts to compensate for each other's weaknesses, effectively mitigating issues such as slow convergence, high intervention frequency, and limited robustness. The primary contributions of this work are summarized as follows:
\begin{enumerate}
\item We propose a dynamic scheduling mechanism within the MoE framework that coordinates IL and RL experts based on action variance. By adjusting the expert allocation according to specific task demands, this approach improves overall system robustness.
\item The proposed pipeline integrates offline pre-training with online fine-tuning, using a small set of demonstration data for expert initialization. This approach enhances co-training efficiency and speeds up convergence.
\item We regularize the RL policy with the action distribution from IL to maintain safety during exploration and improve sample efficiency, thereby reducing the need for human intervention during training.
\end{enumerate}

Evaluation on four real-world tasks shows that the proposed method, after 2 to 5 hours of fine-tuning, achieves an average success rate of 97.5\%. The approach reduces human intervention to an average of 7\% of the total dataset, converges 21\% faster than baseline algorithms, and increases the number of autonomous successful samples by 111.7\%. These results suggest the framework can adapt to various operational requirements, including storage, high-precision assembly, and flexible object manipulation. The dynamic expert fusion strategy offers a practical solution for balancing efficiency and robustness in manipulation tasks, providing a path for integrating IL and RL.

\section{RELATED WORK}

\subsection{IL and RL Fusion Paradigms}

The integration of IL and RL has emerged as a central approach in robotic manipulation, combining the sample efficiency of IL with the environmental adaptability of RL. Existing methodologies typically follow several structural paradigms. One common strategy uses IL for pre-training to establish an initial policy, which addresses the inefficiencies of random exploration in subsequent RL stages \cite{lei2026rl100performantroboticmanipulation}. Other approaches use RL to fine-tune VLA models \cite{chen2026pitextttrlonlinerlfinetuning,doi:10.36227/techrxiv.176531955.54563920/v1}, specifically to mitigate compounding errors and distribution shift. Parallel integration schemes \cite{chen2025conrftreinforcedfinetuningmethod} instead balance imitation and exploration through the simultaneous optimization of multiple objectives. Furthermore, some methods reformulate IL as a reward-shaping task within the RL framework \cite{pmlr-v48-finn16}. More recently, hierarchical and modular designs, such as residual fusion \cite{ankile2025residualoffpolicyrlfinetuning}, freeze a pre-trained BC policy to maintain base capabilities while training a lightweight RL module to provide corrective adjustments. Many contemporary studies combine these paradigms to further enhance manipulation performance.

\subsection{Mixture of Experts in Robotics}

The MoE framework decomposes complex tasks into specialized sub-tasks, facilitating its increasing adoption in robotics. Shazeer et al. \cite{shazeer2017outrageouslylargeneuralnetworks} established the foundation for MoE architectures, which have since been applied to multi-skill learning for quadrupedal robots \cite{10801816, 11246585} to execute dynamic maneuvers through various expert combinations. Through sparse activation and dynamic routing, MoE systems improve policy decision-making efficiency, making them suitable for scenarios requiring real-time responses, such as microsurgery \cite{guo2026moeactscalingmultitaskbimanual}. The scalability and interpretability of this architecture further enable skill decomposition in long-horizon robotic manipulation \cite{cheng2025moedpmoeenhanceddiffusionpolicy}. MoE models also support unified modeling for multi-modal and multi-task learning. For example, integrating force sensing into VLA models \cite{yu2025forcevlaenhancingvlamodels} improves precision manipulation and physical adaptation in contact-rich environments. Despite these advancements, limited research explores applying MoE architectures to the scheduling of heterogeneous experts across IL and RL. The proposed method uses an MoE architecture to achieve dynamic scheduling between IL and RL experts. By employing a gating mechanism that synergistically evaluates motion certainty and contact complexity based on the variance of expert actions, the proposed method achieves expert specialization and dynamic adaptation across diverse task scenarios.

\subsection{Fine-Grained Manipulation in the Real World}

Current policy modeling paradigms for fine-grained manipulation generally fall into three distinct categories. IL \cite{zhao2023learningfinegrainedbimanualmanipulation, black2026pi0visionlanguageactionflowmodel} and offline RL \cite{peng2019advantageweightedregressionsimplescalable} utilize human demonstrations for direct deployment, yet these approaches often struggle with distribution shifts in real-world environments. Alternatively, simulation-based RL leverages domain randomization to bridge the sim-to-real gap, ensuring robustness during real-world deployment. While successful for highly dynamic locomotion tasks \cite{siekmann2021blindbipedalstairtraversal, tan2018simtoreallearningagilelocomotion}, these methods often struggle with precision manipulation because of the difficulty in modeling complex robot-environment interactions. The third approach involves direct real-world training via IL \cite{8968287} or RL \cite{10610040}, which learns physical dynamics without explicit modeling. However, this strategy carries risks of hardware damage and typically requires human intervention \cite{8793698} to maintain safety and facilitate convergence. Modern frameworks increasingly integrate offline pre-training or simulated data into real-world learning schemes. In this study, we employ a framework that combines offline pre-training with online real-world learning, seamlessly and effectively bridging the gap between static datasets and complex, non-stationary real-world environments.

\section{METHOD}

\begin{figure*}[htbp]
	\centering
	\includegraphics[width=1.0\textwidth]{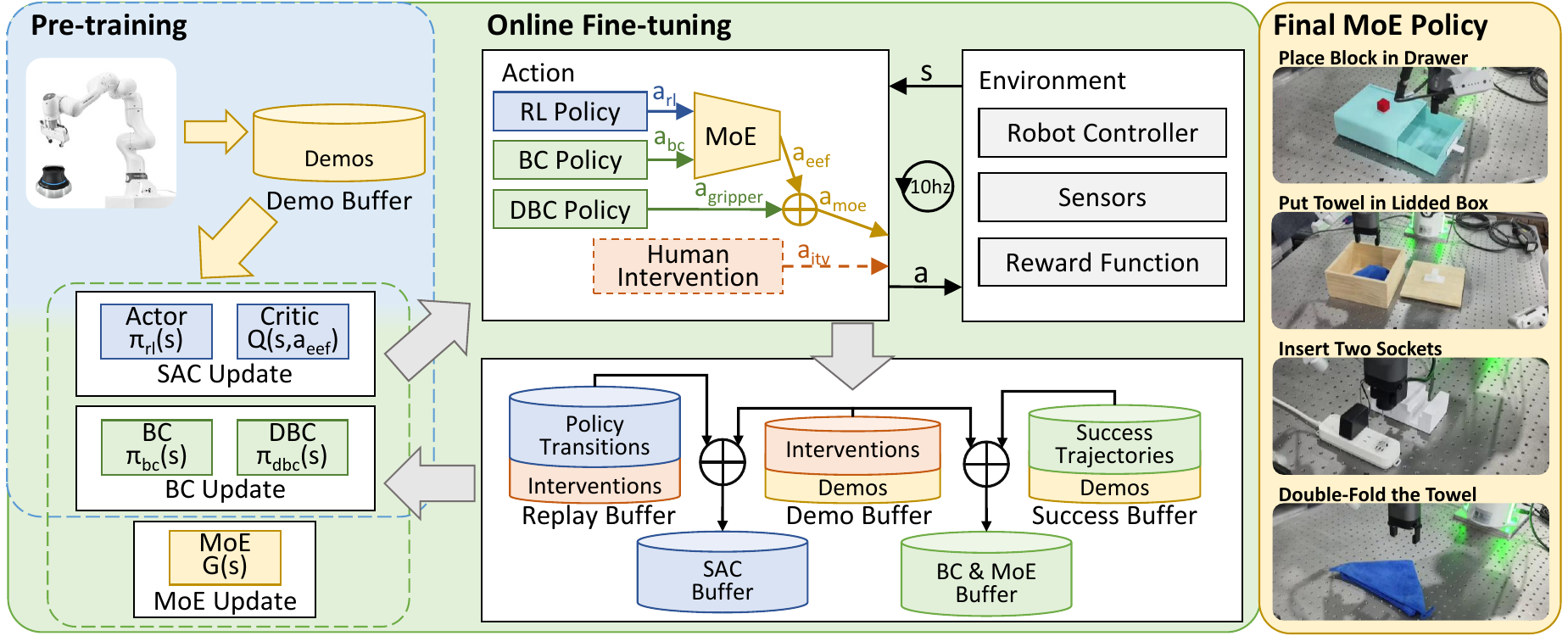}
	\caption{Overview of the MoRI framework. The architecture integrates IL and RL within a MoE system. During offline pre-training, both BC and RL experts are initialized using demonstration data. In the online fine-tuning stage, a gating network manages action selection while a data routing mechanism enables iterative policy refinement for all experts.}
	\label{fig:systemoverview}
\end{figure*}

This study focuses on fine-grained manipulation tasks involving dense contact, formulated as a Markov Decision Process $M=(S, A, T, r, \gamma, \rho_{0})$. The state space $S$ integrates multi-modal sensor data to represent the robot and its environment, while the action space $A$ consists of continuous control commands for the robotic end-effector. Given the inherent dynamics and uncertainty of contact-rich manipulation, the transition function $T$ is highly non-linear, making explicit modeling difficult. The reward function $r$ provides feedback based on task completion, with $\gamma$ serving as the discount factor and $\rho_{0}$ representing the initial state distribution. The objective is to learn an optimal policy $\pi^*:S \to A$ that maximizes the expected cumulative reward $\mathbb{E}[\sum_{t=0}^\infty \gamma^t r(s_t,a_t)]$.

The proposed framework integrates BC and RL through a gating network conditioned on expert action variance. By combining these paradigms, this hybrid approach mitigates the compounding errors of imitation while overcoming the prohibitive cost of blind exploration. As shown in Fig.~\ref{fig:systemoverview}, the system operates in two stages, consisting of offline pre-training followed by online fine-tuning.

\subsection{Stage I: Advantage-Weighted Offline Pre-training}

To address the challenges of inefficient exploration in RL and the limited robustness of IL in complex manipulation tasks, we pre-train our model using a dataset of 20 demonstration samples. In this stage, the expert demonstration buffer $\mathcal{D}_{demo}$ is used to perform supervised fine-tuning on the BC policy and to initialize the RL policy through an AWAC \cite{nair2021awacacceleratingonlinereinforcement} objective. This initialization aligns both IL and RL policies with human strategies, which reduces the cost of trial and error while mitigating slow convergence and distribution shift during online interaction. The MoE component remains untrained during this period to avoid overfitting to the limited demonstration data.

As the network responsible for coarse motions, the BC policy $\pi_{BC}(a|s)$ is fine-tuned on the demonstration dataset $\mathcal{D}_{demo}$ using a standard Mean Squared Error (MSE) loss to establish a motion baseline for subsequent online learning. The loss function is defined as:
\begin{equation}
\mathcal{L}_{BC}(\theta_{bc}) = \mathbb{E}_{(s,a) \sim \mathcal{D}_{demo}} \left[ \left\|\pi_{\theta_{bc}}(s) - a\right\|_{2}^{2} \right]
\end{equation}	
where $\mathcal{L}_{BC}(\theta_{bc})$ is the objective parameterized by $\theta_{bc}$, $\pi_{\theta_{bc}}(s)$ is the predicted action, and $a$ denotes the target end-effector pose increment.

Offline RL often suffers from excessive conservatism, yet stochastic exploration in real-world environments remains prohibitively expensive. We address these issues using AWAC algorithm designed to balance demonstration knowledge with exploratory capabilities. During $N_{offline}$ iterations, the critic and actor networks are updated using transition data sampled from $\mathcal{D}_{demo}$. The critic network is optimized by minimizing the following loss:
\begin{equation}
\mathcal{L}_{Q}(\phi) = \mathbb{E}_{(s,a,s^{\prime},r) \sim \mathcal{D}_{demo}} \left[ \left( Q_{\phi}(s,a) - y \right)^2 \right]
\end{equation}
where the target value $y$ is defined as $y = r(s,a) + \gamma \mathbb{E}_{a^{\prime} \sim \pi_{\theta_{rl}}(a \mid s)} \left[ Q_{\bar{\phi}}(s^{\prime},a^{\prime}) \right]$. We employ clipped double-Q learning, setting $Q_{\bar{\phi}} = \min(Q_{\bar{\phi}_1}, Q_{\bar{\phi}_2})$, to mitigate overestimation bias. The actor network is then updated via advantage weighting:
\begin{equation}
\mathcal{L}_{RL}(\theta_{rl}) = -{\mathbb{E}}_{(s,a) \sim \mathcal{D}_{demo}}\left[\log\pi_{\theta_{rl}}(a|s) \exp\left(\frac{1}{\lambda}A(s,a)\right)\right]
\end{equation}
where $A(s,a)$ is the advantage function and $\lambda$ is a hyperparameter that scales the weighting intensity.

In discrete gripper control, training policies with Deep Q-Network often results in slow convergence for long-horizon tasks. We address this by introducing a discrete BC loss to accelerate the learning process. This loss is defined using a cross-entropy objective:
\begin{equation}
\mathcal{L}_{DBC}(\theta_{dbc}) = -\mathbb{E}_{(s, a_{g}) \sim \mathcal{D}_{demo}} \left[ \log \pi_{\theta_{dbc}} (a_{g}|s) \right]
\end{equation}
where the gripper action $a_g$ comprises three discrete modes: open, hold, and closed. By providing direct supervision over the discrete action space, this BC component avoids the stochasticity and moving-target issues inherent in bootstrapping-based target network updates and leads to faster convergence than standard value-based methods.

\subsection{Stage II: Mixture-of-Experts Driven Online Fine-tuning}

\begin{figure*}[!htbp]
	\centering
	\includegraphics[width=1.0\textwidth]{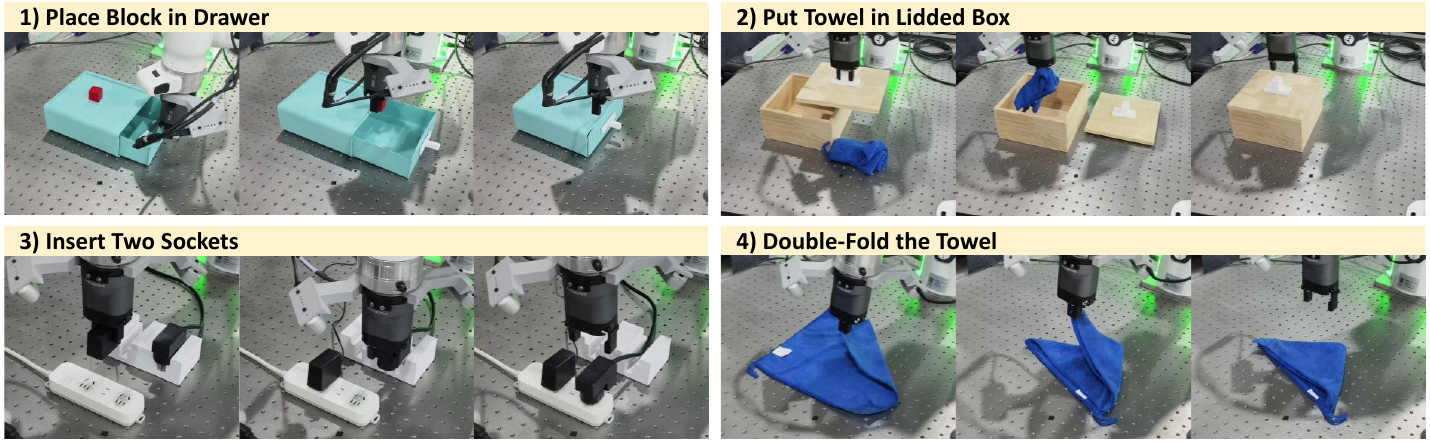}
	\caption{Overview of real-world experimental tasks, including 1) place block in drawer, 2) put towel in lidded box, 3) insert two sockets, and 4) double-fold the towel.}
	\label{fig:experimentoverview}
\end{figure*}

The online stage maintains three evolving datasets to support optimization guided by the MoE. The expert demonstration buffer $\mathcal{D}_{demo}$ is initialized with offline data and updated with human interventions collected during online execution. Similarly, the success buffer $\mathcal{D}_{success}$ builds upon $\mathcal{D}_{demo}$ by incorporating successful trajectories from online training, which helps the system account for the variability of real-world tasks. The replay buffer $\mathcal{D}_{replay}$ begins as an empty set and stores real-time interaction data, where the data distribution is shaped by both MoE-selected actions and human corrections to cover a broad range of scenarios. These datasets provide the foundation for BC, Discrete BC, and Soft Actor-Critic (SAC) training. Through selective uniform sampling, they also facilitate the refinement of the MoE, ensuring that the specific data requirements of each expert are satisfied.

The online fine-tuning stage employs task-specific loss functions for the BC, SAC, and MoE modules to optimize the policy across the three datasets. The loss for the BC expert $\mathcal{L}_{BC}(\theta_{bc})$ and the discrete BC loss for the gripper $\mathcal{L}_{DBC}(\theta_{dbc})$ remain consistent with the offline pre-training stage, with the training data replaced by $\mathcal{D}_{bc}$, which is constructed by uniformly sampling from both $\mathcal{D}_{success}$ and $\mathcal{D}_{demo}$.

The SAC framework optimizes the agent through a combination of critic and actor updates. The critic loss ensures stable value estimation by minimizing the MSE between the predicted and target Q-values:
\begin{equation}
    \begin{aligned}
        \mathcal{L}_{Q}(\phi)&=\mathbb{E}_{(s, a, s^{\prime}, r) \sim \mathcal{D}_{rl}}\Big[ Q_{\phi}(s, a) - \\
        &\quad \left( r + \gamma \mathbb{E}_{a^{\prime} \sim \pi_{\theta_{rl}}(a \mid s)}\left[Q_{\bar{\phi}}\left(s^{\prime}, a^{\prime}\right)\right] \right) \Big]^2
    \end{aligned}
\end{equation}
where $Q_{\bar{\phi}}$ is the target Q-network used to mitigate overestimation bias. Training samples in $\mathcal{D}_{rl}$ are sampled with equal probability from $\mathcal{D}_{replay}$ and $\mathcal{D}_{demo}$ separately. The corresponding actor loss balances Q-value maximization, entropy regularization for exploration, and a regularization relative to the BC policy:
\begin{equation}
    \begin{aligned}
        \mathcal{L}_{RL}(\theta_{rl})&=-\mathbb{E}_{s \sim \mathcal{D}_{rl}}\left[\mathbb{E}_{a \sim \pi_{\theta_{rl}}(a \mid s)}\left[Q_{\phi}(s, a)\right] \right. \\
        &\quad + \alpha \mathcal{H}\left(\pi_{\theta_{rl}}(\cdot \mid s)\right) \\
        &\quad \left. - \beta \left\|\pi_{\theta_{rl}}(s) - \pi_{\theta_{bc}}(s)\right\|_2^2\right]
    \end{aligned}
\end{equation}
where the policy entropy $\mathcal{H}(\pi{\theta_{rl}}(\cdot \mid s))$ encourages exploration and prevents premature convergence, with the adaptive temperature $\alpha$ controlling the exploration strength. The final MSE term provides policy regularization, ensuring that the RL agent's exploration remains within the distributional support of the BC policy. The coefficient $\beta$ modulates the strength of this constraint.

For a given state $s$, the gating network $g_{\psi}(s)$ generates weights $w_{bc}$ and $w_{rl}$ subject to $w_{bc} + w_{rl} = 1$, representing the contributions of the BC and RL experts, respectively. The gating network within the MoE framework is optimized via a multi-component loss function $\mathcal{L}_{MoE}(\psi)$. Instead of a standard MSE, this objective combines weighted variance minimization, load balancing, expert specialization, and entropy:
\begin{equation}
    \begin{aligned}
        \mathcal{L}_{MoE}(\psi) &=
        \mathbb{E}_{s \sim \mathcal{D}_{bc},[w_{bc}, w_{rl}] \sim g_{\psi}(s)}\bigg[ \\
        &\quad \underbrace{w_{bc} \cdot \sigma_{bc}(s) + w_{rl} \cdot \sigma_{rl}(s)}_{\text{Weighted Variance Loss}} \\
        &\quad + \underbrace{\alpha \cdot (0.5 - |w_{bc} - 0.5|)}_{\text{Specialization Loss}} \\
        &\quad + \underbrace{\beta \cdot [\left(\mathbb{E}[w_{bc}] - 0.5\right)^2 + \left(\mathbb{E}[w_{rl}] - 0.5\right)^2]}_{\text{Load Balancing Loss}} \\
        &\quad + \underbrace{ \gamma \cdot \left(-\mathcal{H}\left(g_{\psi}(\cdot \mid s)\right)\right)}_{\text{Entropy Regularization}}\bigg]
    \end{aligned}
\end{equation}
where the terms $\sigma_{bc}(s)$ and $\sigma_{rl}(s)$ denote the variance of the action outputs from each expert. While $\mathcal{D}_{bc}$ corresponds to the dataset used by the BC expert, the hyperparameters $\alpha, \beta$, and $\gamma$ control the trade-off between the different objectives. These components guide expert routing by balancing exploration and exploitation, equalizing expert utilization, and preventing premature convergence. Specifically, the framework uses the variance of action outputs as a proxy for expert confidence. Although variance scales often differ across experts, the specialization and load balancing terms remove the need for direct variance comparison. 

Following the generation of expert weights, the MoRI strategy employs a hard selection mechanism at each time step to determine the action $a$. Specifically, the system selects the output of the BC expert $\pi_{\theta_{bc}}(s)$ if $w_{bc} > w_{rl}$, and otherwise defaults to the RL expert $\pi_{\theta_{rl}}(s)$:
\begin{equation}
a =
\begin{cases}
    \pi_{\theta_{bc}}(s), & \text{if } w_{bc} > w_{rl}, \\
    \pi_{\theta_{rl}}(s), & \text{otherwise}.
\end{cases}
\end{equation}
This action is combined with the discrete gripper command $a_g$ from the DBC expert to form the final control output.

\begin{table*}[h]
\centering
\caption{Performance Comparison: ConRFT\cite{chen2025conrftreinforcedfinetuningmethod} vs. MoRI}
\begin{adjustbox}{max width=\linewidth}
\begin{tabular}{@{}rccccccccccccccc@{}}
\toprule
\multirow{2}{*}{\textbf{\makecell{\\Task}}} &  
\multicolumn{2}{c}{\textbf{\makecell{Training Time (mins)}}} & 
\multicolumn{2}{c}{\textbf{\makecell{Success Rate (\%)}}} &
\multicolumn{2}{c}{\textbf{\makecell{Episode Length}}} &
\multicolumn{2}{c}{\textbf{\makecell{Demo / Replay\\Buffer Size Ratio (\%)}$\downarrow$}} &
\multicolumn{2}{c}{\textbf{\makecell{Auto-Success / Replay\\Buffer Size Ratio (\%)}$\uparrow$}} \\
\cmidrule(lr){2-3} \cmidrule(lr){4-5} \cmidrule(lr){6-7} \cmidrule(lr){8-9} \cmidrule(lr){10-11}
& \makecell{ConRFT} & \makecell{MoRI}
& \makecell{ConRFT} & \makecell{MoRI}
& \makecell{ConRFT} & \makecell{MoRI}
& \makecell{ConRFT} & \makecell{MoRI}
& \makecell{ConRFT} & \makecell{MoRI}  \\
\midrule
Place Block in Drawer
& 257   & \textbf{142}\makecell[t]{(-44.8\%)}
& 75    & \textbf{100}
& \textbf{93.1}  & 125.2
& 38.1  & \textbf{5.5}\makecell[t]{(-85.6\%)}
& 49.0  & \textbf{85.4}\makecell[t]{(+74.3\%)}   \\
Put Towel in Lidded Box
& 250   & \textbf{194}\makecell[t]{(-22.4\%)}
& 0     & \textbf{95}
& -     & \textbf{169.1}
& 73.3  & \textbf{7.0}\makecell[t]{(-90.5\%)}
& 18.8   & \textbf{81.3}\makecell[t]{(+332.5\%)}   \\
Insert Two Sockets
& 252   & \textbf{241}\makecell[t]{(-4.4\%)}
& 80    & \textbf{95}
& \textbf{97.3}  & 180.2
& 25.5  & \textbf{6.2}\makecell[t]{(-75.7\%)}
& 51.9  & \textbf{74.3}\makecell[t]{(+43.2\%)}   \\
Double-Fold the Towel
& 336   & \textbf{288}\makecell[t]{(-14.3\%)}
& 75    & \textbf{100}
& \textbf{78.4}  & 171.2
& 61.3  & \textbf{9.4}\makecell[t]{(-84.7\%)}
& 28.3  & \textbf{72.3}\makecell[t]{(+155.5\%)}   \\
\midrule
\textbf{Average}
& 273.8     & \textbf{216.3}\makecell[t]{(-21.0\%)}
& 57.5      & \textbf{97.5}
& \textbf{89.6}      & 161.4
& 49.6      & \textbf{7.0}\makecell[t]{(-85.8\%)}
& 37.0      & \textbf{78.3}\makecell[t]{(+111.7\%)}   \\
\bottomrule
\multicolumn{12}{p{\linewidth}}{\footnotesize Autonomous success data is defined as expert policy data extracted from successful trajectories, excluding the data of human intervention.}
\end{tabular}
\end{adjustbox}
\label{tab:rl_performance_comparison}
\end{table*}

\section{EXPERIMENTS AND RESULTS}
\subsection{Overview of Experiments}

We assess performance across four representative tasks requiring both coarse motion planning and fine-grained contact control, as shown in Fig.~\ref{fig:experimentoverview}. The specific configurations for each task are as follows:
\begin{enumerate}
\item \textbf{Place Block in Drawer}: The agent opens a drawer, grasps a block, places it inside, and closes the drawer. The block is initialized at a random orientation within a 5 $\text{cm} \times $ 5 $\text{cm}$ area above the table-mounted drawer.
\item \textbf{Put Towel in Lidded Box}: This task involves lifting a lid, depositing a towel, and replacing the lid. The towel is placed within a 3 $\text{cm} \times $ 3 $\text{cm}$ random region on the table, while the box is fixed to the table.
\item \textbf{Insert Two Sockets}: The agent inserts two-pin and three-pin plugs into precise slots. While the sockets and holders are fixed to the table, the plugs are stored with a 1 $\text{cm}$ directional random error.
\item \textbf{Double-Fold the Towel}: This task requires double folding a towel through multi-contact force control. The towel is laid flat with a fixed orientation and a 2 $\text{cm}$ random error in the $xy$ plane.
\end{enumerate}

Experiments are conducted on a Franka Research 3 robotic arm platform using visual observations from three cameras, which are cropped and resized to a $128 \times 128 \times 3$ resolution. For most tasks, the camera configuration includes two wrist-mounted units and one global side-view camera, whereas the plug-insertion task utilizes one wrist camera and two side-view cameras. The system records synchronized multi-modal sensor data, encompassing end-effector poses, velocities, and contact forces. The action space for real-world execution is seven-dimensional, representing six-dimensional incremental end-effector poses and a binary gripper command. Both data collection and policy execution operate at 10Hz. Models are implemented in JAX and trained on a single NVIDIA RTX 4090 GPU, with policy networks based on a multilayer perceptron architecture. Before online training, a human operator provides success and failure demonstrations to train a binary classifier for reward feedback. Additionally, 20 human demonstrations are collected for offline pre-training.

The evaluation protocol is designed to address the following research questions:
\begin{enumerate}
\item[\textbf{Q1.}]Does MoRI enhance manipulation performance relative to the baseline RL model?
\item[\textbf{Q2.}]Can MoRI outperform its individual sub-experts?
\item[\textbf{Q3.}]In what manner does MoRI schedule different experts?
\item[\textbf{Q4.}]How does BC regularization influence performance and the smoothness of expert switching?
\item[\textbf{Q5.}]How does the performance of the gripper expert differ between DQN and DBC?
\end{enumerate}

\begin{figure*}[htbp]
	\centering
	\includegraphics[width=1.0\textwidth]{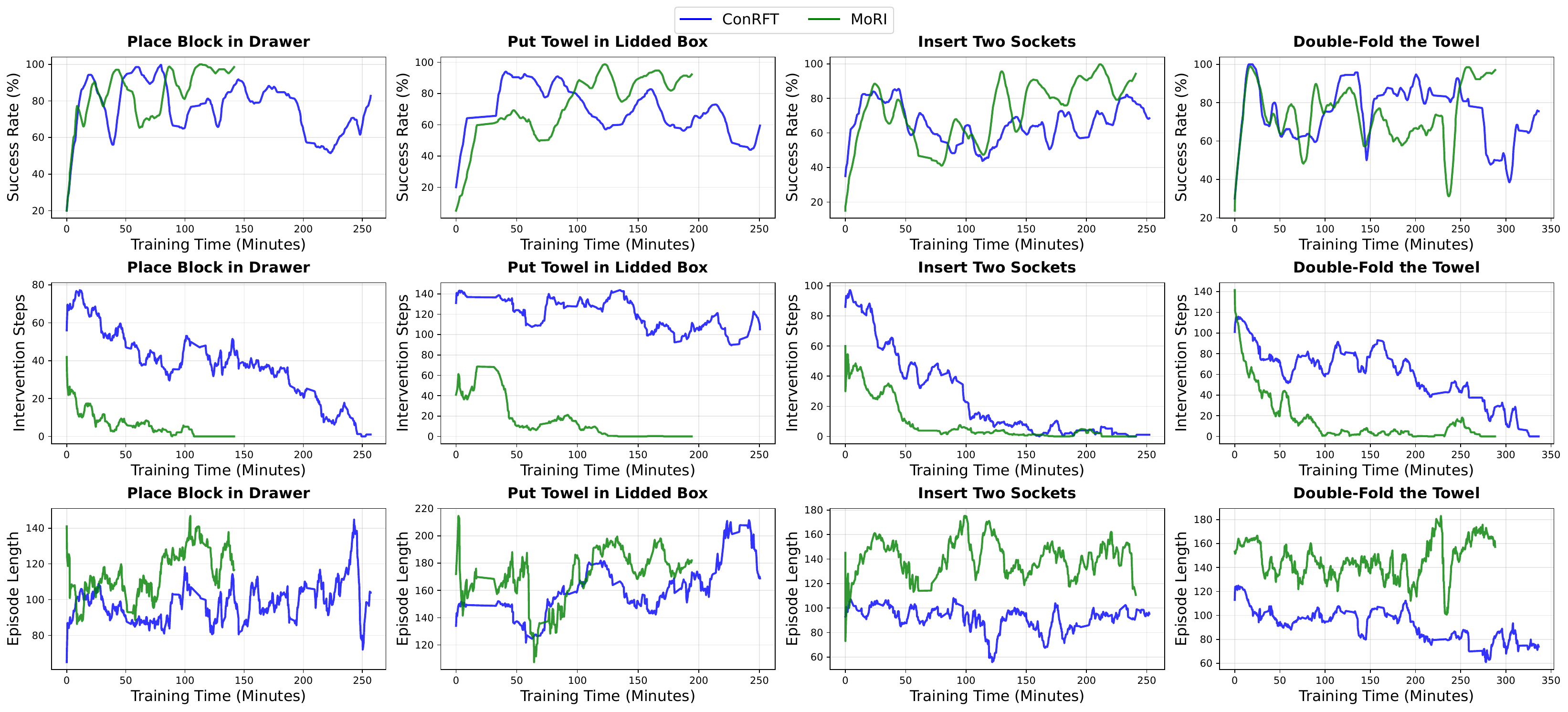}
	\caption{Learning curves during online training. Each panel compares the performance of ConRFT\cite{chen2025conrftreinforcedfinetuningmethod} and MoRI in terms of success rate, intervention rate, and episode length. Data points are calculated using a moving average over 20 episodes. Recorded success rates may reflect the combined effect of the policy and human assistance, potentially exceeding the standalone capability of the autonomous agent.}
	\label{fig:experimentcurvefigure}
\end{figure*}

\subsection{Experimental Results}

We evaluate the proposed method across several tasks, with results summarized in Fig.~\ref{fig:experimentcurvefigure} and Table~\ref{tab:rl_performance_comparison}. Performance is measured using the success rate, training duration, average episode length, and buffer efficiency, with success rate and episode length averaged over 20 independent trials. 

\textbf{Q1: Does MoRI enhance manipulation performance relative to the baseline RL model?} MoRI outperforms ConRFT\cite{chen2025conrftreinforcedfinetuningmethod} by maintaining BC regularization through independent BC experts and a hybrid scheduling strategy. In contrast, ConRFT relies on a weighted fusion strategy where the BC loss constraint diminishes rapidly during online training, often leading to a loss of policy guidance. As shown in Table~\ref{tab:rl_performance_comparison}, our method reduces average training time by 21.0\%, demonstrating higher sample efficiency. MoRI maintains success rates above 95.0\% across all tasks, exceeding the 57.5\% success rate of ConRFT. These results suggest robust error recovery in contact-rich scenarios. Furthermore, MoRI improves data utilization while reducing reliance on human demonstrations. The results in Table~\ref{tab:rl_performance_comparison} show that the proportion of demonstration data in the MoRI replay buffer is 7.0\%, significantly lower than the 49.6\% required by ConRFT, even as the proportion of successful trajectories increases to 78.3\%. Intervention curves in Fig.~\ref{fig:experimentcurvefigure} further indicate that MoRI converges faster with fewer total interventions, suggesting it utilizes high-quality experience more efficiently to achieve optimized policies.

\begin{table}[htbp]
\centering
\caption{Performance Comparison: MoRI vs. Individual Sub-Experts}
\begin{adjustbox}{max width=\linewidth}
\begin{tabular}{@{}rcccccc@{}}
\toprule
\multirow{2}{*}{\textbf{\makecell{\\Task}}} 
& \multicolumn{3}{c}{\textbf{Success Rate (\%)}} 
& \multicolumn{3}{c}{\textbf{Episode Length}} \\
\cmidrule(lr){2-4} \cmidrule(lr){5-7}
& BC & RL & MoRI & BC & RL & MoRI \\
\midrule
Place Block in Drawer        
& 80    & 60     	& \textbf{100}   
& 136.0 & 149.5     & \textbf{125.2} \\
Put Towel in Lidded Box  
& 60    & 75     	& \textbf{95}    
& 254.0 & 204.1     & \textbf{169.1} \\
Insert Two Sockets          
& 10    & 90     	& \textbf{95}    
& 228.0 & 201.9     & \textbf{180.2} \\
Double-Fold the Towel       
& 0     & 85     	& \textbf{100}   
& -     & \textbf{144.2}     & 171.2 \\
\midrule
\textbf{Average}            
& 37.5  & 77.5   	& \textbf{97.5}  
& 206.0   & 174.9   & \textbf{161.4} \\
\bottomrule
\end{tabular}
\end{adjustbox}
\label{tab:experts_ablation}
\end{table}

\textbf{Q2: Can MoRI outperform its individual sub-experts?} To investigate whether MoRI schedules sub-experts appropriately, we compare its performance against BC and RL baselines across four physical manipulation tasks. As shown in Table~\ref{tab:experts_ablation}, MoRI consistently outperforms individual sub-experts. Specifically, MoRI achieves an average success rate of 97.5\%, a marked improvement over BC (37.5\%) and RL (77.5\%). This performance gain extends to efficiency, where MoRI requires an average of 161.4 steps, compared to 206.0 steps for BC and 174.9 steps for RL. Despite the fluctuating performance of BC and RL across the four tasks, MoRI remains consistently effective. These results suggest that MoRI develops an adaptive scheduling scheme during training tailored to different tasks. By leveraging the complementary strengths of RL and BC, MoRI effectively mitigates their respective limitations, achieving superior performance in complex robotic manipulation.

\begin{figure*}[!htbp]
	\centering
	\includegraphics[width=1.0\textwidth]{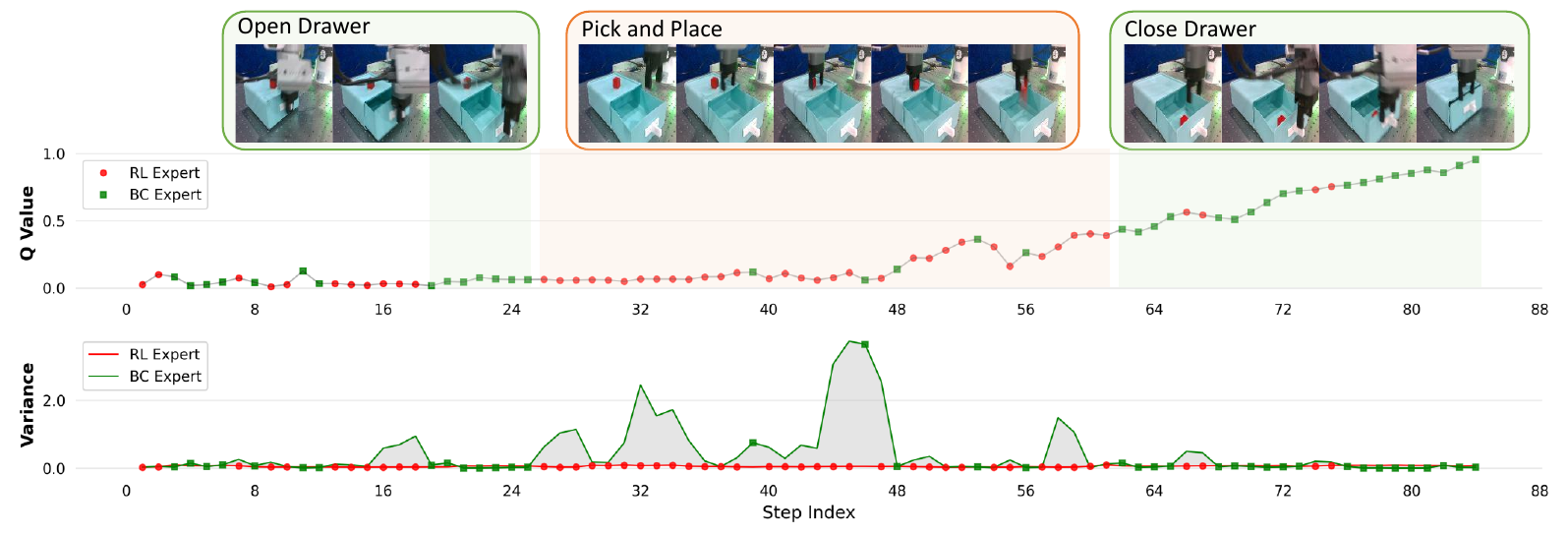}
	\caption{The evolution of Q-values and variance for RL and BC experts within the MoRI framework.}
	\label{fig:qablationfigurereal}
\end{figure*}

\begin{figure}[!htbp]
	\centering
	\includegraphics[width=0.45\textwidth]{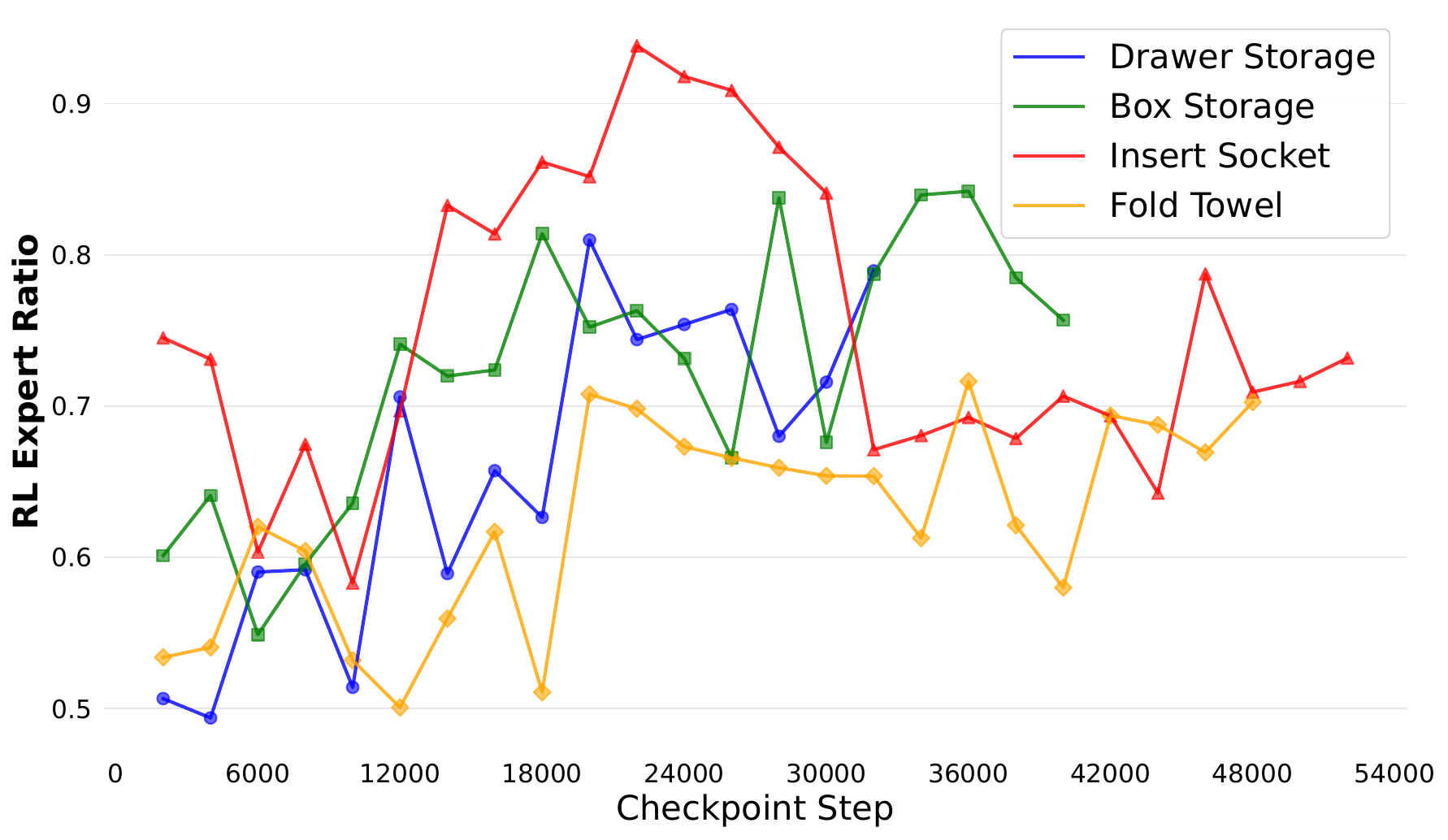}
	\caption{Evolution of the RL expert selection ratio over training steps in MoRI, averaged across five trajectories sampled from each corresponding checkpoint.}
	\label{fig:vablationfigurereal}
\end{figure}

\textbf{Q3: In what manner does MoRI schedule different experts?} Taking the drawer storage task as an example, Fig.~\ref{fig:qablationfigurereal} demonstrates that MoRI typically selects the BC expert for relatively fixed subtasks, such as opening or closing drawers. These actions exhibit high consistency and low fitting variance, making them well-suited for the BC policy. In contrast, fine-grained manipulation tasks like picking and placing are more sensitive to target randomness, in which the system relies more heavily on the RL expert, benefiting from dense Q-value guidance and maintaining lower variance. As shown in Fig.~\ref{fig:vablationfigurereal}, the proportion of RL expert calls increases from 50\%--60\% to 70\%--80\% during training. This shift arises because MoRI's load balancing is anchored to the dataset distribution rather than the online state-action distribution. Initially, the minimal gap between these distributions ensures a balanced deployment of experts, harmonizing BC's conservatism with RL's exploration. However, as RL-driven exploration and human interventions cause the action distribution to evolve, the static dataset begins to lag behind. Meanwhile, the RL expert, bolstered by dense Q-network guidance from accumulated data, exhibits superior adaptability to the current distribution, enabling a seamless transition from conservative imitation to efficient, exploration-driven RL.

\begin{table}[!htbp]
\centering
\caption{Ablation: MoRI Regarding BC Regularization}		
\begin{adjustbox}{max width=\linewidth}
\begin{tabular}{@{}rcccccc@{}}
\toprule
\textbf{Drawer Task} &
\textbf{\makecell{Training\\Time (mins)}} & 
\textbf{\makecell{Success\\Rate (\%)}} &
\textbf{\makecell{Episode\\Length}} &
\textbf{\makecell{Demo\\Ratio (\%)}$\downarrow$} &
\textbf{\makecell{Auto-Success\\Ratio (\%)}$\uparrow$} \\
\midrule
Base Policy
& 142    	& 100     		& 125.2 
& 5.5 		& 85.4     \\
w/o BC Regularization 
& 221    	& 70     			& 122.1  
& 45.4 		& 41.9     \\
\bottomrule
\multicolumn{6}{p{\linewidth}}{\footnotesize Demo Ratio represents Demo / Replay Buffer Size Ratio.} \\
\multicolumn{6}{p{\linewidth}}{\footnotesize Auto-Success Ratio represents Auto-Success / Replay Buffer Size Ratio.}
\end{tabular}
\end{adjustbox}
\label{tab:experts_bc_ablation}
\end{table}

\begin{figure}[!htbp]
	\centering
	\includegraphics[width=0.45\textwidth]{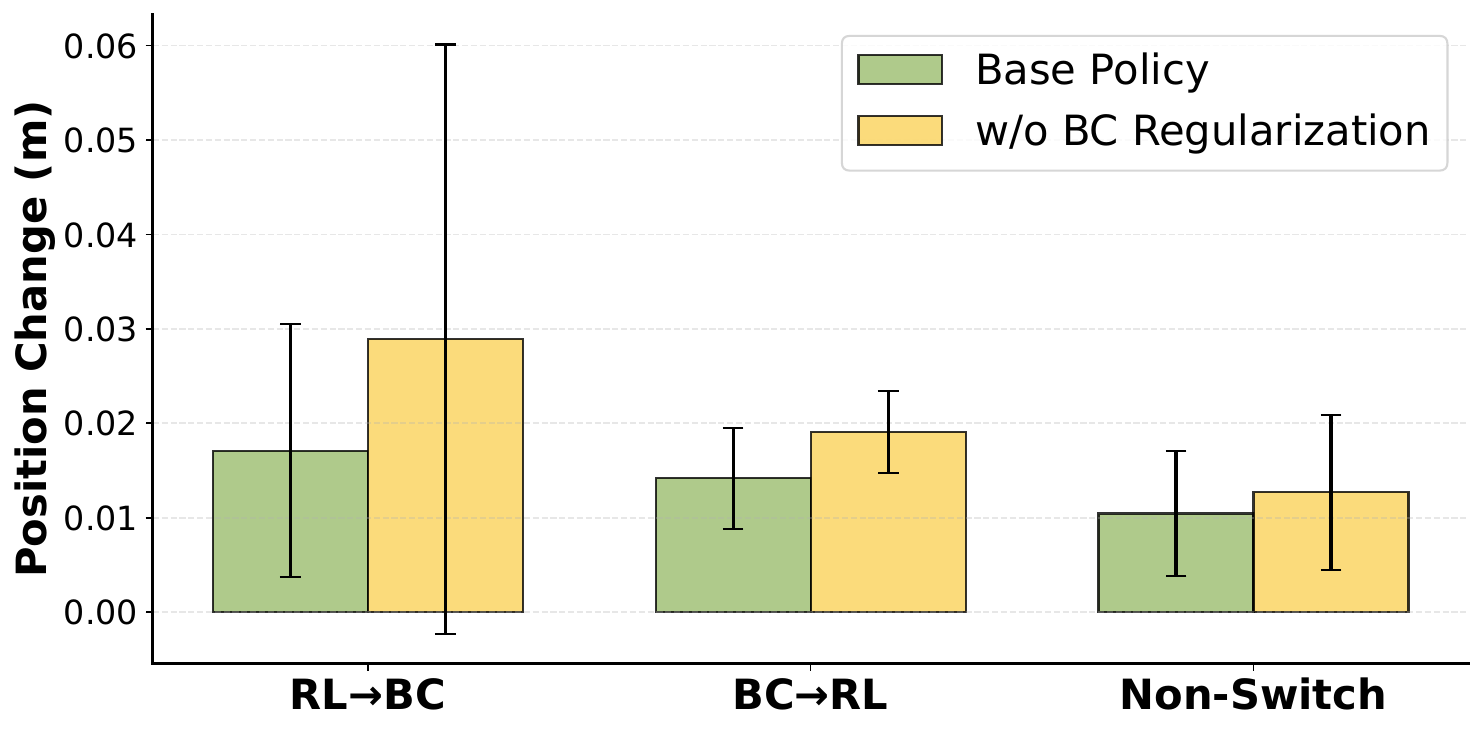}
	\caption{Action fluctuations during expert switching within the MoRI framework.}
	\label{fig:switchablationfigurereal}
\end{figure}

\textbf{Q4: How does BC regularization influence performance and the smoothness of expert switching?} Results in Table~\ref{tab:experts_bc_ablation} indicate that human intervention in the drawer storage task rises from 5.5\% to 45.4\% upon removing the BC regularization, highlighting that the absence of this constraint leads the RL expert to engage in excessive stochastic exploration. This absence also extends the training duration from 142 minutes to 221 minutes, while the success rate falls to 70\% and the autonomous success ratio drops from 85.4\% to 41.9\%. These experiments confirm that the regularization restricts the search space to reasonable regions, mitigating the slow convergence typical of RL training. As shown in Fig.~\ref{fig:switchablationfigurereal}, base policy maintains motion continuity during dynamic scheduling. Consistent with normal execution, the end-effector position varies by only about 0.017 m during transitions between heterogeneous experts, confirming high action-level alignment within the framework. Conversely, removing the BC regularization significantly increases position shifts at the moment of switching, with jumps exceeding 0.029 m during the RL-to-BC transition, coupled with a high standard deviation of 0.031 m. These findings confirm that BC regularization reduces disparities in action distributions, ensuring smooth motion during hard switching.

\textbf{Q5: How does the performance of the gripper expert differ between DQN and DBC?} Comparing Table~\ref{tab:rl_performance_comparison} and Table~\ref{tab:experts_gripper_ablation} shows that DBC substantially outperforms DQN when integrated as the gripper expert. While DQN yields a slightly shorter average episode length, its success rate falls to 72.5\%, significantly lower than the 97.5\% achieved by DBC. Additionally, the DQN model requires a 23.4\% intervention rate to maintain a 60.0\% autonomous success rate and takes 323.3 minutes to converge, trailing DBC in both efficiency and reliability. For long-horizon tasks involving six or more gripper switches, such as placing items into lidded boxes, the DQN expert fails to converge. These results demonstrate that DBC is more robust for gripper control in complex manipulation sequences.

\begin{table}[!htbp]
\centering
\caption{Performance Results using DQN instead of DBC}		
\begin{adjustbox}{max width=\linewidth}
\begin{tabular}{@{}rcccccc@{}}
\toprule
\textbf{Task} &  
\textbf{\makecell{Training\\Time (mins)}} & 
\textbf{\makecell{Success\\Rate (\%)}} &
\textbf{\makecell{Episode\\Length}} &
\textbf{\makecell{Demo\\Ratio (\%)}$\downarrow$} &
\textbf{\makecell{Auto-Success\\Ratio (\%)}$\uparrow$} \\
\midrule
Place Block in Drawer   
& 210    & 90     		& 104.0  
& 13.8 & 71.4     \\
Put Towel in Lidded Box
& 422    & 30     		& 185.2   
& 30.6 & 46.4     \\
Insert Two Sockets         
& 381    & 75     		& 135.1  
& 25.8 & 57.3    \\
Double-Fold the Towel   
& 280     & 95     		& 103.9
& 23.4     & 65.0     \\
\midrule
\textbf{Average}            
& \textbf{323.3}  & \textbf{72.5}   			& \textbf{132.1}  
& \textbf{23.4}   & \textbf{60.0}    \\
\bottomrule
\end{tabular}
\end{adjustbox}
\label{tab:experts_gripper_ablation}
\end{table}

\section{CONCLUSION}
This paper presents MoRI, a framework that integrates RL and IL using a MoE architecture. Through a two-stage pipeline consisting of offline pre-training and online fine-tuning, MoRI employs a gating mechanism conditioned on expert action variance to manage scheduling. The system assigns deterministic coarse motions to the IL expert, while reserving fine manipulations with higher uncertainty for the RL expert. By using IL to provide regularization for the RL component, the framework improves both exploration safety and sample efficiency. Evaluations on four complex real-world manipulation tasks show that MoRI achieves an average success rate of 97.5\% within 2 to 5 hours of online fine-tuning. Compared to the ConRFT baseline, this approach reduces training convergence time by 21\% and decreases human intervention by 85.8\%. These results suggest that MoRI effectively synthesizes IL and RL, offering a reliable method to balance policy efficiency and stability in manipulation.

\textbf{Limitations and Future Work.} Despite its robustness, the MoRI framework faces limitations in generalization and autonomous recovery under significant perturbations. To address these, the current dual-expert architecture could evolve into a hierarchical system incorporating Vision-Language Models for high-level reasoning and zero-shot transfer. Furthermore, to mitigate the intensive human labor and operational costs of real-world training, future work will explore World Models to partially replace physical trials through simulated imagination. Finally, integrating online continual learning and sparse expert activation will be essential for managing long-horizon tasks and achieving more autonomous, large-scale deployment.

\bibliographystyle{IEEEtran}
\bibliography{references}
\end{document}